\documentclass[10pt,twocolumn,letterpaper]{article}

\usepackage{iccv}              % To produce the CAMERA-READY version
\definecolor{iccvblue}{rgb}{0.21,0.49,0.74}
\usepackage[pagebackref,breaklinks,colorlinks,allcolors=iccvblue]{hyperref}

\usepackage{bbm}
\usepackage{multirow}
\usepackage{ulem}
\usepackage{makecell} 
\usepackage{diagbox}
\usepackage{graphicx}
\usepackage{amsmath}
\usepackage{amssymb}
\usepackage{booktabs}

\def\paperID{12055} % *** Enter the Paper ID here
\def\confName{ICCV}
\def\confYear{2025}

\title{Proactive Scene Decomposition and Reconstruction}

\author{
    Baicheng Li$^{1,2}$ \qquad
    Zike Yan$^{3}$ \qquad
    Dong Wu$^{1,2}$ \qquad
    Hongbin Zha$^{1,2}$$^\dag$
    \\[1.5ex]
    $^{1}$School of Intelligence Science and Technology, Peking University \\
    $^{2}$National Key Laboratory of General Artificial Intelligence \qquad $^{3}$AIR, Tsinghua University
    \\[1.5ex]
    {\tt\small libc@pku.edu.cn \quad yanzike@air.tsinghua.edu.cn \quad riserwu@stu.pku.edu.cn \quad zha@cis.pku.edu.cn}
}

\begin{document}

\twocolumn[{%
	\renewcommand\twocolumn[1][]{#1}%
	\maketitle
    \vspace{-9mm}
	\begin{center}
		\centering
		\includegraphics[width=0.93\linewidth]{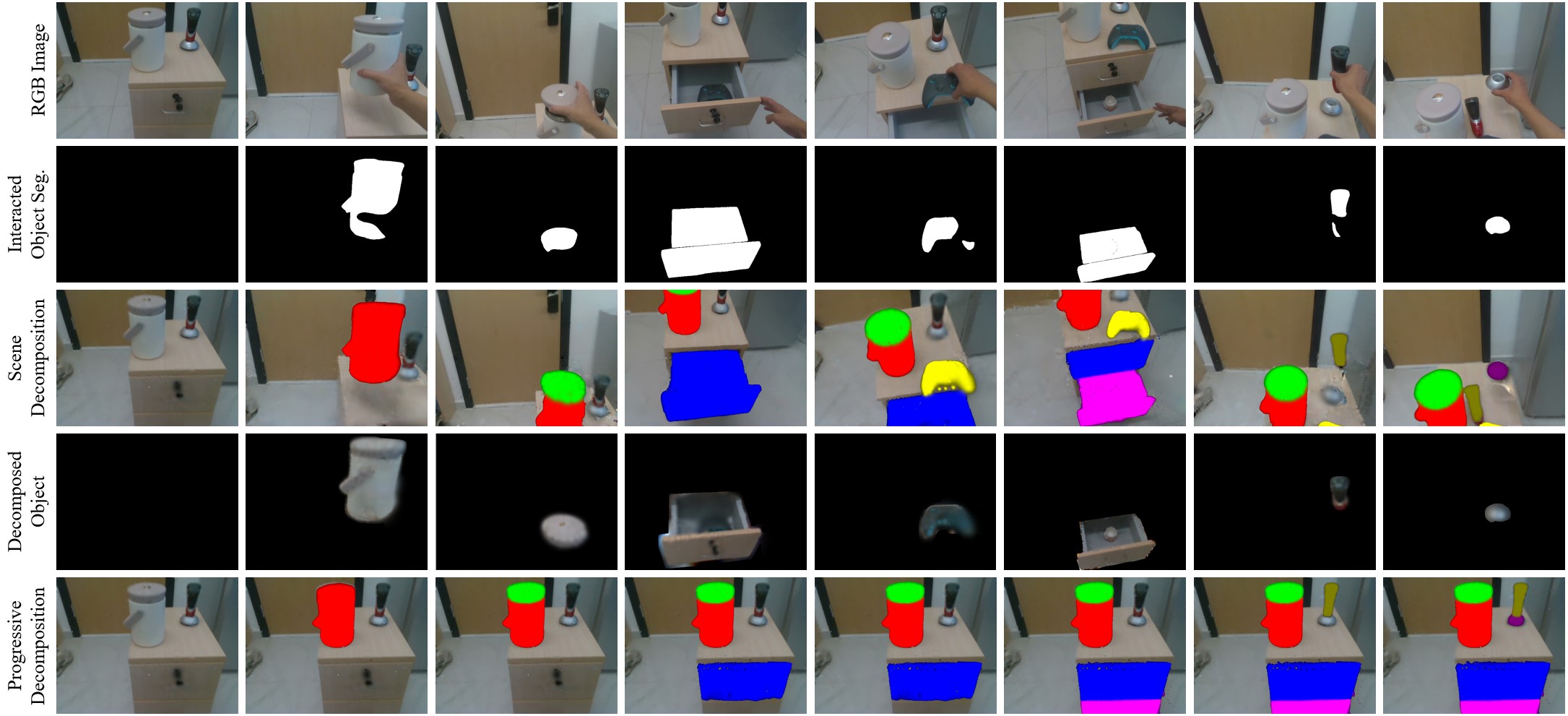}
		\captionof{figure}{We introduce a dynamic SLAM system to tackle the proactive scene decomposition and reconstruction from egocentric live streams (first row). Only objects that are under proactive interactions will be decomposed (second) to maintain a consistent granularity. This fashion leads to photorealistic modeling of the environment (third to fifth rows), enabling progressive scene decomposition (fifth row) and robust object tracking (third and fourth rows).}
		\label{fig:cover}
	\end{center}%
}]

\begin{abstract}
\vspace{-10mm}

\noindent Human behaviors are the major causes of scene dynamics and inherently contain rich cues regarding the dynamics. This paper formalizes a new task of proactive scene decomposition and reconstruction, an online approach that leverages human-object interactions to iteratively disassemble and reconstruct the environment. By observing these intentional interactions, we can dynamically refine the decomposition and reconstruction process, addressing inherent ambiguities in static object-level reconstruction. The proposed system effectively integrates multiple tasks in dynamic environments such as accurate camera and object pose estimation, instance decomposition, and online map updating, capitalizing on cues from human-object interactions in egocentric live streams for a flexible, progressive alternative to conventional object-level reconstruction methods. Aided by the Gaussian splatting technique, accurate and consistent dynamic scene modeling is achieved with photorealistic and efficient rendering. The efficacy is validated in multiple real-world scenarios with promising advantages.
\end{abstract}  
\vspace{-6mm}
\section{Introduction}
\label{sec:intro}

Understanding the ever-changing environment is vital but fundamentally challenging for the vision and robotics communities. While many existing methods attempt to solve this problem by breaking down complex environments into manageable and semantically meaningful components, they often rely on passive data acquisition and pre-defined models to tackle 4D reconstruction or dynamic SLAM problems. While these methods can be effective in certain scenarios, they often struggle to capture the true dynamic nature of the environments, resulting in incomplete or inaccurate models. A key issue is the lack of consideration for human activity, which is often the dominant force in shaping the dynamics of most real-world environments. Consequently, the methods miss out on valuable contextual insights derived from human-object interactions, which can provide critical context for understanding spatiotemporal relationships in a dynamic setting. By leveraging cues from these interactions, we can iteratively decompose independently moving regions, creating a more flexible and adaptive approach to modeling dynamic environments.

In this paper, we formalize a new task of \textit{proactive scene decomposition and reconstruction}: an online process of dynamically disassembling and reassembling the environment given observations of ongoing human-object interactions.
Traditional object-level scene reconstruction methods~\cite{Siddiqui2023cvpr,Bhalgat2024nips,chen2023arxiv,Liu2024cvpr,Ye2024eccv} have primarily focused on static environments, where the major challenge lies in the inherent ambiguity of decomposition. Besides, the intersected areas between objects lead to incomplete observations. Attempts have been made to either enforce consistency across views~\cite{Siddiqui2023cvpr, Bhalgat2024nips} or perform inpainting for surface completion~\cite{Mirzaei2023cvpr,Zhong2024eccv}. However, as illustrated in~\cref{fig:cover}, both the completion and the static decomposition are ill-posed. For instance, should we separate the drawer from the cabinet, or treat the contents of the drawer as part of the cabinet? What does the inside view look like if the drawer is closed during the data capture?  We argue that effective decomposition should not be static, as the granularity is highly context-dependent. Instead, the decomposition process should be progressive and guided by interaction. Unlike 4D reconstruction or dynamic SLAM, which aim to handle arbitrary scene dynamics, we restrict the task to recovering a compositional scene representation from first-person live streams under intentional interactions, allowing for more controllable scene decomposition and accurate reconstruction by best exploiting the interaction cues. 

Simultaneously addressing both scene decomposition and reconstruction in dynamic environments leads to a complicated system with the need for accurate camera pose estimation, object pose estimation, instance decomposition, and the fusion of past observations into a globally consistent map. The integration of these modules is usually difficult, as they are unstable and sensitive to outliers. However, as we demonstrate in the following section, hand-object interactions offer a stable and controllable definition of the compositional granularity as the individually moving part, allowing for progressively accurate object masks to be generated in a dynamic context. With this approach, the pose estimation task, both for the camera and the objects, becomes simplified, effectively reducing them to locally static problems that are trivial to solve, leading to accurate and holistic modeling of the environment.

For the proposed new task, we also introduce an online algorithm. Compared to offline methods, the online approach enables users to receive timely feedback, providing guidance during capturing and laying the foundation for incremental map updates. For streaming inputs, our method performs online camera pose tracking, object pose tracking, scene decomposition, and reconstruction. We fully leverage the interaction information available in egocentric inputs to achieve a high-quality object-level decomposed map reconstruction. To summarize, our main contributions include:
\begin{itemize}
\item We introduce a new task of proactive scene decomposition and reconstruction, aiming to decompose and reconstruct the environment based on online human-object interactions, offering a flexible alternative to conventional object-level reconstruction methods, allowing adaptive and progressive processes in response to the interaction cues.
\item We propose an online dynamic SLAM system for proactive scene decomposition. Guided by the interaction priors, our system achieves more accurate scene decomposition, pose estimation, and reconstruction in an online fashion.
\item We effectively combine the temporal constraints of foundation models and spatiotemporal consistency for modeling scene dynamics. The integration of both constraints along with fixed granularity induced by interactions enables our algorithm to achieve promising local homogeneity.
\end{itemize}

\section{Related Work}
\label{sec:related_work}
\subsection{Object-decomposed radiance fields}
Recent advances in radiance fields have garnered widespread attention due to the photorealistic rendering results. As a global representation, decomposing the radiance field into individual components is one natural extension for downstream tasks that require local editing and reasoning, such as scene editing~\cite{Zhong2024eccv} and realistic simulation~\cite{Yang2023cvpr,Torne2024rss}. 
~\cite{Ost2021cvpr} introduces a neural rendering technique and decomposes dynamic scenes with scene graphs.~\cite{Yang2021iccv} designs a novel two-pathway architecture, where the scene branch encodes the geometry and appearance of the background, and the object branch encodes prior-conditioned learnable representations. ObjectSDF~\cite{Wu2022eccv} and ObjectSDF++~\cite{Wu2023iccv} establish a connection between the semantics of each object and the corresponding geometry, enabling the creation of object-compositional neural implicit surfaces guided by RGB images and their corresponding instance masks. However, these methods generally require ground-truth instance masks and object association information as inputs. 

To address the object decomposition problem, Panoptic Lifting~\cite{Siddiqui2023cvpr} and Contrastive Lifting~\cite{Bhalgat2024nips} adopt the linear assignment and contrastive learning to achieve object separation in 3D radiance fields given image segmentation predictions across views. With the emergence of Segment Anything (SAM)~\cite{sam} and the video segmentation models like SAM2~\cite{sam2}, the training of object-level radiance fields can be supervised directly from the predicted masks~\cite{chen2023arxiv,Liu2024cvpr,Ye2024eccv}. However, these methods often encounter issues due to ambiguous segmentation granularity. D2NeRF~\cite{Wu2022nips} and NeuralDiff~\cite{Tschernezki20213dv}] attempt to decouple dynamic scenes through a simple motion segmentation, which can be defined precisely as the moving part of the environment. Additionally, some methods~\cite{yang2024deformable,wu20244d} focus on modeling dense deformation fields to capture the spatiotemporal information of the scene. However, most existing methods neglect the strong cues inherited in the interaction between agent and environment. We share a similar motivation with a concurrent work of EgoGaussian~\cite{Zhang2025_3dv}, which leverages the hand-object interaction in egocentric videos for spatial-temporal modeling of dynamic environments and tracking rigid object motion. In contrast to the offline optimization process, our method extends the paradigm further by exploiting the instant feedback and temporal continuity within the streaming data to enable progressive scene decomposition and online holistic reconstruction of the dynamic environment.

\subsection{Agent-in-the-loop scene understanding}
Besides the passive scene understanding, agent-in-the-loop exploits the agent engagement to actively perceive and analyze the environment. For instance, Roboexp~\cite{Jiang2024corl} introduces action-conditioned scene graphs, where robots accumulate information through active interactions to capture the geometry and the structure of the surroundings. Similarly, Nagarajan and Grauman~\cite{Nagarajan2020nips} introduce affordance landscapes, enabling robots to learn about the actions that can be performed within a 3D environment. The approach helps robots recognize the potential for interaction with novel objects and enhance their ability to adapt to new environments. The cluttered environment always poses challenges for object recognition and segmentation. In~\cite{Wu2022eccv}, robust object recognition is achieved by combining perception with interaction. A similar system is adopted in Autoscanning~\cite{Xu2015tog} to couple scene reconstruction with proactive object analysis. In~\cite{Mishra2009iccv}, scene segmentation is improved by selectively attending to certain areas, highlighting the role of fixation in active segmentation. Recent work has also delved into uncertainty-aware segmentation. Yu and Choi~\cite{Yu2022eccv} propose a self-supervised method for interactive object segmentation through singulation-and-grasping approach. This demonstrates how robots can learn from their interactions without requiring explicit supervision, a significant step toward autonomous scene understanding. Fang et al.~\cite{Fang2024iros} explore how robots reduce the uncertainty of object segmentation through physical actions. This concept is also adopted in RISEG~\cite{Qian2024icra} by exploiting body frame-invariant features and robot interaction to correct inaccurate segmentation.

We refer readers to~\cite{Bohg2017tro}, a comprehensive review on interactive perception. This work sets a foundation for how robots can use their actions to improve scene understanding and vice versa. Besides the robot-in-the-loop scene understanding, human interventions also help to reduce the perception ambiguities. Many works study the visual perception in an egocentric video given hand-object interaction priors~\cite{Zhang2023iccv, Cheng2023nips,Narasimhaswamy2024cvpr}. There are also studies working on 3D object decomposition through live annotations. iLabel~\cite{Zhi2022ral} exploits the shared embedding space of jointly optimized neural fields, enabling efficient scene labeling given sparse clicks. Similarly, in Total-decom~\cite{Lyu2024cvpr}, extensive involvement of human labeling is reduced to enforce real-time control of quality and granularity of the scene decomposition with minimal interaction. 
These works collectively contribute to the advancement of agent-in-the-loop scene understanding, while we take a step further to directly perform online scene reconstruction and decomposition given an egocentric live stream of hand-object interaction. The scene is progressively decomposed and reconstructed in a unified SLAM system to jointly optimize scene radiance, camera motion, object poses, and instance segmentation.

\subsection{Dynamic SLAM}
The presence of dynamic objects introduces significant challenges to camera tracking and mapping as common consistency across views is assumed under static scenarios. The aim of dynamic SLAM is to remove features that violate the cross-view consistency constraints, ensuring precise camera tracking and reliable static map reconstruction. Relevant methods are commonly divided into two categories. The first approach utilizes warping or reprojection, as in~\cite{cheng2019improving, scona2018staticfusion, palazzolo2019refusion}, to detect inconsistencies in visual appearance or spatial geometry, thereby identifying dynamic regions in images. The second approach~\cite{he2017mask} leverages prior knowledge, such as semantic categories, to determine whether an object is dynamic. Some methods further combine these two approaches. DynaSLAM~\cite{bescos2018dynaslam} and DRG-SLAM~\cite{wang2022drg} remove features that belong to pre-defined categories or bypass geometric constraints. In contrast, SLAMANTIC~\cite{schorghuber2019slamantic} and CFP-SLAM~\cite{hu2022cfp} use projection to verify observations within pre-defined categories, selectively removing only those features that exhibit inconsistencies.

\begin{figure*}[ht]
    \centering
    \includegraphics[width=\textwidth]{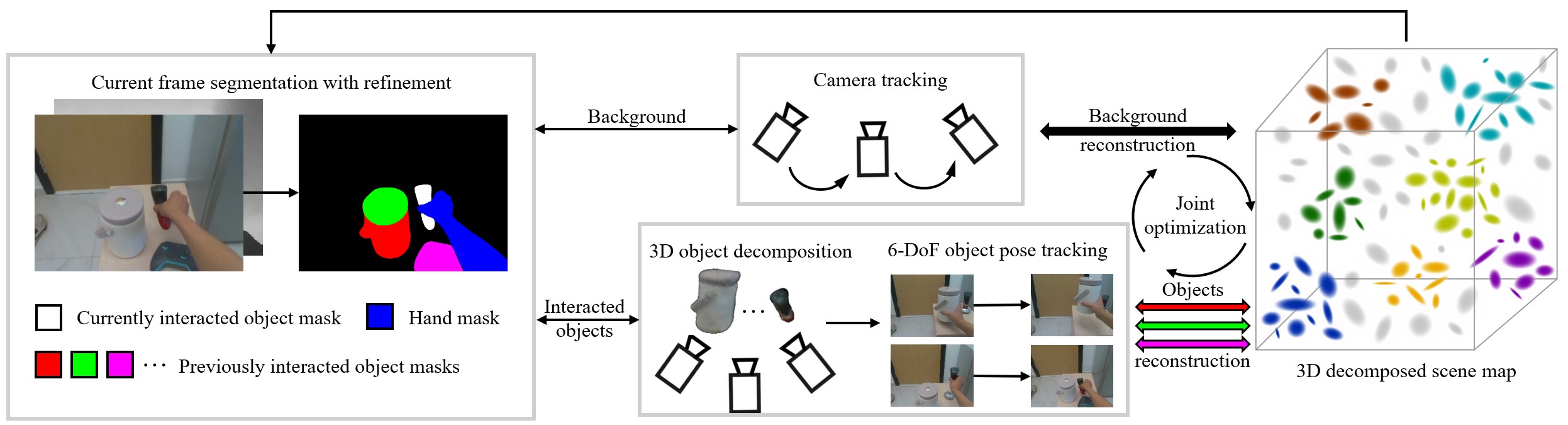}
    \caption{Overview of our method. With well-defined decomposition granularity induced by motion, the online system achieves reliable camera tracking and scene reconstruction, allowing progressive decomposition and robust instance tracking.}
    \label{fig:pipeline}
\end{figure*}

Our method can also be seen as a combination of these two approaches, leveraging both inconsistencies between observations and the map, and prior knowledge from user interactions to identify dynamic objects. However, unlike these existing methods, which primarily focus on reconstructing the static part of the environment, our approach not only reconstructs the static background but also decouples and reconstructs all interacted objects. This allows us to obtain a more informative and holistic understanding of the dynamic environment.

\section{Overview}
\label{sec:overview}
We aim to achieve online scene decomposition and reconstruction from egocentric RGB-D videos. The system takes streaming observations as inputs, where hand-object interactions are proactively carried out. The proposed system assumes that all objects moved by interactions exhibit approximately rigid body motion. The primary output is a scene representation that consists of background areas $G_B$ along with the decomposed instances $G_{O_i}$ ($i = 1, \dots, n$) as $\{G_B, G_{O_1}, \dots, G_{O_n}\}$. The decomposition is carried out progressively, where a new instance will be initialized once the hand-object interaction is identified.

In practice, we maintain a Gaussian-based representation for better photorealism. To ensure fast convergence and prevent overfitting during the online optimization, the Gaussian primitive is parameterized by RGB color $\mathbf{c} \in \mathbb{R}^3$, center position $\mu \in \mathbb{R}^3$, isotropic variance $r \in \mathbb{R}$, and opacity $o \in \mathbb{R}$. Each object, once decomposed, is tracked and reconstructed independently. View-dependent color, depth, silhouette images and instance segmentation can be rendered using the Gaussian splatting technique as follows:
\begin{equation}
    \hat{\mathbf{C}}[u, v] = \sum_{i \in N} \mathbf{c}_i f_i[u, v] \prod_{j=1}^{i-1} \left( 1 - f_j[u, v] \right),
\end{equation}
\begin{equation}
    \hat{\mathbf{D}}[u, v] = \sum_{i \in N} d_i f_i[u, v] \prod_{j=1}^{i-1} \left( 1 - f_j[u, v] \right),
    \label{eq:depth_render}
\end{equation}
\begin{equation}
    \hat{\mathbf{S}}[u, v] = \sum_{i \in N} f_i[u, v] \prod_{j=1}^{i-1} \left( 1 - f_j[u, v] \right),
    \label{eq:silhouette_render}
\end{equation}
\begin{equation}
    \hat{\mathbf{I}}[u, v] = \sum_{i \in N} k_i f_i[u, v] \prod_{j=1}^{i-1} \left( 1 - f_j[u, v] \right),
\end{equation} 
where $k_i$ is the ID of the decomposed instance $G_{O_i}$ the Gaussian belongs, $f_i[u, v]$ is computed as:
\begin{equation}
    f[u, v] = o \exp \left( -\frac{\|[u, v] - \mu_{2D}\|^2}{2r_{2D}^2} \right),
\end{equation}
\begin{equation}
    \mu_{\text{2D}} = K \frac{\tilde{E_t} \mu}{d}, \quad r_{\text{2D}} = \frac{f r}{d}, \quad d = (\tilde{E_t} \mu)_z,
\end{equation}
where $\Tilde{E_t}$ represents the relative pose of the corresponding object with respect to the camera at time $t$.

The overarching goal of the scene decomposition and reconstruction is the joint optimization of camera pose, object pose, Gaussian parameters, and the assignment of instance labels:

\begin{equation}
    L = \lambda_{p} L_p + \lambda_{d} L_d + \lambda_{ID} L_{ID},
    \label{eq:global}
\end{equation}
where $L_p, L_d, L_{ID}$ are the expected L1 loss of color, depth, and instance segmentation given pixels within the mask M.
% \begin{equation}
%     L_{p}=\frac{1}{|M|} \sum_{(u, v) \in M} \left| C \left[u, v \right]-\hat{C} \left[u, v \right]\right|,
% \end{equation}
% \begin{equation}
%     L_{d}=\frac{1}{|M|} \sum_{(u, v) \in M} \left| D \left[u, v \right]-\hat{D} \left[u, v \right]\right|,
% \end{equation}
% \begin{equation}
%     L_{ID}=\frac{1}{|M|} \sum_{(u, v) \in M} \left| I \left[u, v \right]-\hat{I} \left[u, v \right]\right|,
% \end{equation}

% Pixels with silhouette values below the threshold $\lambda_S$ are considered as unseen areas, thus constructing the mask $M_S$ for camera localization.
% The Gaussian splatting technique is employed to represent both the background and the decomposed objects in the scene. 

As illustrated in~\cref{fig:pipeline}, the key to the problem, as also indicated in~\cref{eq:global}, is the decomposition that enforces the joint optimization of the map and poses as independent tasks for each decomposed instance under a local static assumption.
We will show as follows that the priors originated from hand-object interactions and the spatiotemporal consistency maintained within the map jointly assures the accurate decoupling of objects from the background progressively, facilitating both pose estimation and scene reconstruction.
\section{The Proactive Mapping System}
\label{sec:system}
\begin{figure}[ht]
    \centering
    \includegraphics[width=0.48\textwidth]{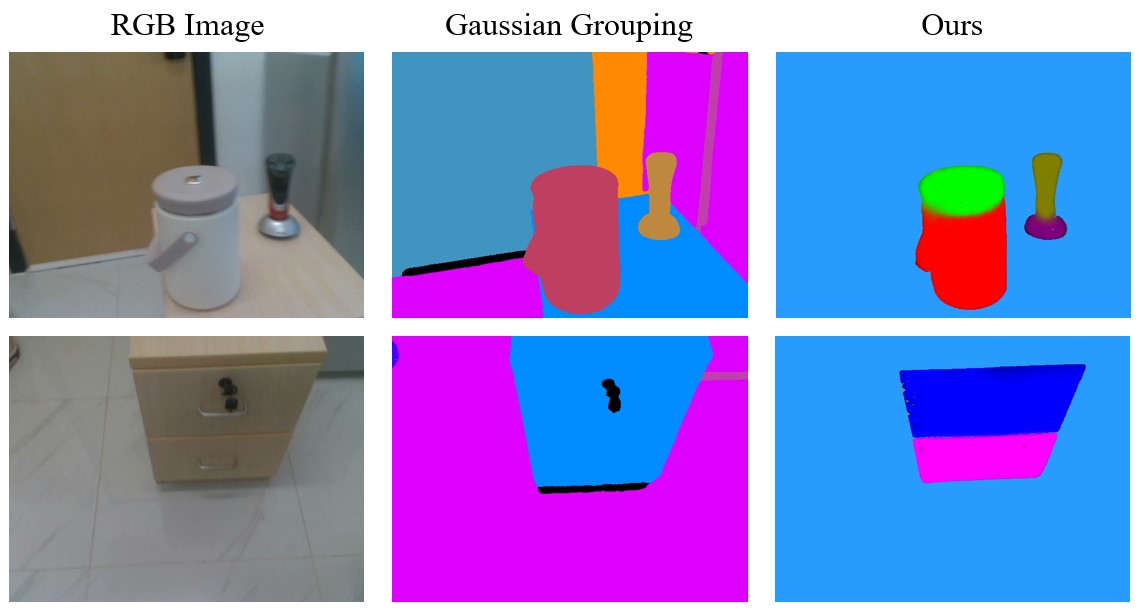}
    \caption{Comparison of scene decomposition with Gaussian Grouping~\cite{Ye2024eccv}. Gaussian Grouping relies on the pre-defined segmentation granularity of the foundation models, whereas our method achieves adaptive decomposition given live interaction.}
    \label{fig:gaussian_grouping}
\end{figure}

We formulate the online scene decomposition and reconstruction under proactive hand-object interactions as an object-decomposed dynamic SLAM problem. The proposed system includes four modules: prompted segmentation, camera and object pose estimation, mask refinement, and decomposed scene reconstruction. The system undergoes iterative optimization that progressively decomposes instances under interactions and updates the locally independent maps.
\subsection{Prompted segmentation}
As clarified above, we aim to maintain a fixed granularity for scene decomposition, defining it as the independently moving part. To achieve this, we extract information from the 3D scene map to determine the prompts for the segmentation module, effectively controlling the decomposition granularity. Specifically, as multi-view consistency only holds under static assumptions, the motion leads to inconsistency between the rendering results from the map and the instant observation. 
Similar to~\cite{li2024learn}, we render a depth map based on the estimated camera pose and compute the differences compared to the observed depth map. Pixels with significant differences are regarded as inconsistent regions. Note that inconsistency may not only be caused by motion, but also by inaccurate pose estimation and map parameters. We adopt a filtering mechanism to divide the image into uniform grids and quantify the proportion of inconsistent pixels within each grid. A grid will be marked as dynamic if the portion exceeds a certain value: 
\begin{equation}
    \frac{\sum_{(u, v) \in S_{grid}} \mathbbm{1}\left(\hat{D}[u,v] - D[u,v] > t_d\right)}{|S_{grid}|} > t_p,
\end{equation}
where $t_d$ and $t_p$ are hyperparameters for thresholding. 

% The center of those identified dynamic grids are point prompts that indicate the moving status and are fed into a SAM2~\cite{sam2} segmentation model. Note that the system is established under a strong assumption that all object motion is caused by proactive interaction, we also utilize the hand position to validate the point prompts. The system detects hand positions with a YOLO model to obtain a box prompt. The output of SAM2 model given the point prompts will be checked to see if the region is adjacent to the hand position. 

Subsequently, we detect connected marked grids to form coherent inconsistent area, then extract its centroid as SAM2 prompt for segmenting the interacted object.  To validate segmentation accuracy, we concurrently execute hand localization via YOLO and SAM2, then verify spatial adjacency between the segmented object mask and hand region in both RGB and depth domains. This dual-space proximity check enforces physical interaction constraints to ensure logical segmentation results.

Note that SAM2 is a video segmentation model, where the mask decoder does not merely take encoded prompt features as input. A memory bank of previous observations is maintained and cross-attended with the prompted features for predicting the segmentation. Therefore, the prompted segmentation will only occur when a new instance undertakes the hand-object interactions for the first time. The following frames will track the activated instance through the cross-attended memory bank. To label the entire instance in the 3D space instead of merely the corresponding Gaussian primitives associated with the current view, we need to propagate the mask back to all past keyframes. We empirically find that the encoded prompt features can be directly utilized for segmenting previous keyframes as they remain temporally consistent across views for the same instance.

\begin{figure*}[ht]
    \centering
    \includegraphics[width=\textwidth]{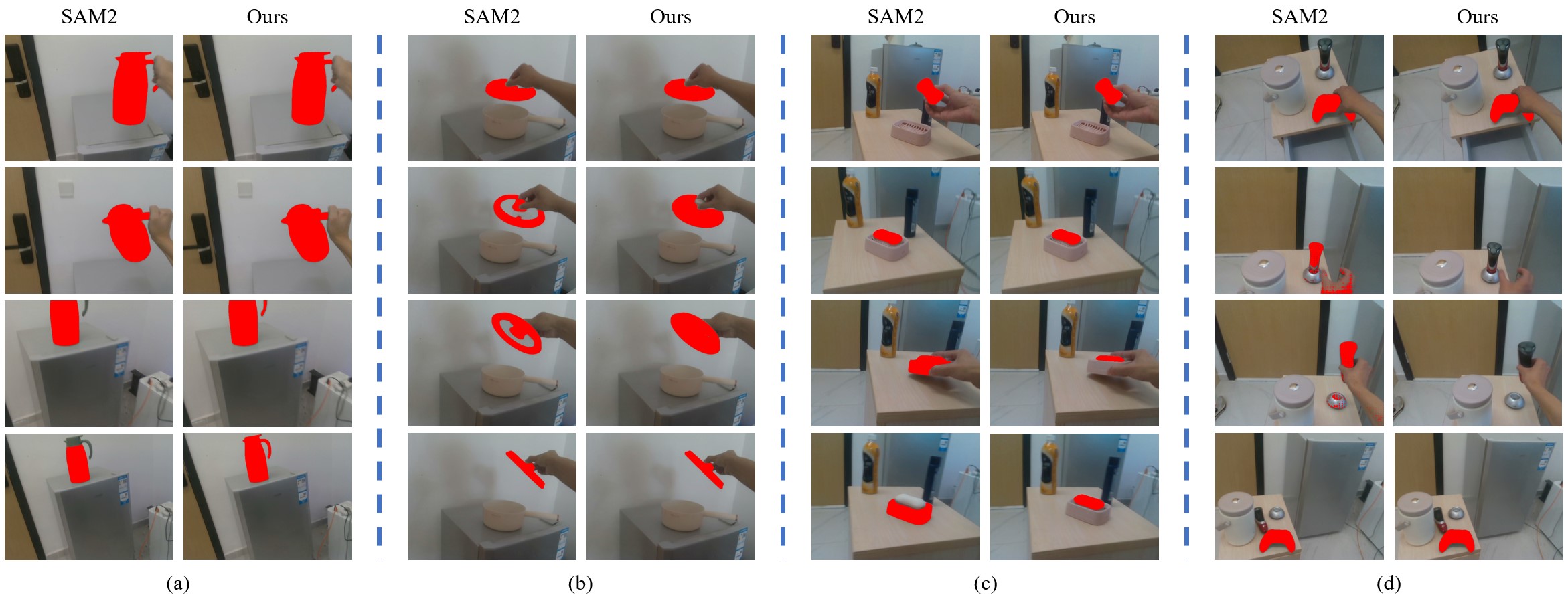}
    \caption{Comparison of segmentation results between SAM2~\cite{sam2} and our method given instances under proactive interactions.}
    \label{fig:segmentation}
\end{figure*}

\subsection{Camera and object pose estimation}
In the online reconstruction setting, for each input frame, we estimate both the camera pose and the poses of all interacted objects, which are prerequisites for global scene reconstruction and object-level refinement. The optimization is performed over rotation and translation parameters, corresponding to the camera and interacted objects, respectively, and is guided by~\cref{eq:global}, with weights $\lambda_p^{ctrack}, \lambda_d^{ctrack}, \lambda_{ID}^{ctrack}$ for camera tracking and $\lambda_p^{otrack}, \lambda_d^{otrack}, \lambda_{ID}^{otrack}$ for object tracking.

A key difference between the two lies in the masking approach. Both employ a silhouette mask $M_S$ to exclude previously unobserved pixels during optimization. However, to mitigate the interference of human and interacted object motion on camera localization, the camera pose estimation further incorporates the previously mentioned human mask $M_h$ and interacted object mask $M_o$ from the current frame, along with the rendered mask $M_{\hat{o}}$ of the interacted object from the scene map.
% In the online reconstruction setting, for each input frame, we first estimate its camera pose. It is optimized using~\cref{eq:global} with weights $\lambda_p^{ctrack}$, $\lambda_d^{ctrack}$ and $\lambda_{ID}^{ctrack}$. During camera motion, some captured pixels may not be included in the existing map. Therefore, during localization, these pixels should be excluded using a masking approach. To achieve this, we render the silhouette of each pixel as~\cref{eq:silhouette_render} and construct the mask $M_S$ with a threshold of 0.99. Additionally, to mitigate the interference of human and interacted object motion on camera tracking, we incorporate the previously mentioned human mask $M_h$ and interacted object mask $M_o$ from the current frame, along with the rendered mask $M_{\hat{o}}$ of the interacted object from the scene map.

% For object tracking, which assumes a rigid body motion, we follow a similar approach. After decoupling the interacted object from the background, performing online 6-DoF pose tracking is a prerequisite for refining its reconstruction. This is optimized using~\cref{eq:global} with weights $\lambda_p^{otrack}$, $\lambda_d^{otrack}$ and $\lambda_{ID}^{otrack}$. Furthermore, we compute the silhouette mask $M_S$ to determine which pixels will be used for back-propagation.

Once object tracking is complete, we follow the Gaussian Splatting-based SLAM approach to densify previously unobserved regions using depth information. For background areas, we directly initialize new Gaussians at the corresponding positions, while for regions of the interacted object, we warp the positions back to their expected locations based on the current estimated object pose.

% We compute the photometric loss $L_p$ and the depth loss $L_d$ between the rendered image and the real image:
% \begin{equation}
%     L^1 = \lambda_{p}^1 L_p + \lambda_{d}^1 L_d.
% \end{equation}
% During camera tracking, we use $M_{\text{track}} = M_S \cup M_h \cup M_o \cup M_{\hat{o}}$ as the mask for dynamic regions, while using other valid pixels to compute the loss and optimize the camera extrinsics.

% After decoupling the interacted object from the background, to further refine its reconstruction results, we first need to perform online 6-DoF pose tracking of the object. We adopt a similar approach to camera tracking to estimate the object pose. At each time step, the object translation and rotation are represented by a 3D vector and a quaternion, both of which are optimizable. The loss is calculated based on the rendered color and depth of the interacted object, and using Eq. 3, we compute the silhouette to determine which pixels will be used for back-propagation to optimize the pose parameters.

% Upon finishing object tracking, we follow the Gaussian Splatting-based SLAM approach to densify previously unobserved regions using depth information. For background areas, we directly initialize new Gaussians at the corresponding positions, while for regions of the interacted object, we warp the positions back to their expected locations based on the current estimated object pose.

\subsection{Segmentation refinement}
\label{subsec:mask_refinement}
Though SAM2 achieves promising results for video segmentation, the spatial consistency is not well exploited due to the image domain inputs. As illustrated in~\cref{fig:segmentation}, we notice typical failure modes during the proactive interactions. Benefiting from the unified framework to keep track of the entire sets of instances within the environment, the dense SLAM system is complementary to handle these failures.

One typical issue is that objects may be partially or fully outside the camera’s field of view due to factors like camera angles or hand occlusion. Thanks to the photorealistic and efficient rendering of Gaussian primitives, we can assign a virtual camera to check if the instance is fully within the field of view. As illustrated in~\cref{fig:segmentation}(a), the kettle lid is erroneously excluded from the segmentation when the kettle reappears in the frame. To address this, we design a flexible-length memory bank to ensure that at least one complete observation of the object is retained in the memory queue.  Based on the current state of the memory bank, we dynamically adjust the length of the memory queue to best retain the most complete observations. This strategy effectively mitigates segmentation errors caused by incomplete or occluded observations in specific periods of the video sequence. 
% Moreover,~\cref{fig:segmentation}(d) demonstrates that when the object is completely absent in certain frames, SAM2 is likely to produce erroneous segmentation results. We simply set the mask to all zeros to avoid excessive addition of prompts for refinement.

Another issue is the inter-frame segmentation inconsistencies. The reliance on 2D information constrains the segmentation accuracy and temporal consistency. As illustrated in~\cref{fig:segmentation}(b,c), we first check whether the previously identified inconsistent area is covered by the predicted mask. We then perform rigid object pose tracking and verify whether the rendered region 
of the interacted object matches the mask. Additional positive or negative point prompts will be added if the mask fails to cover the areas adequately or if it excessively overlaps with the rendered region. The conditions for mask refinement are defined as:
\begin{equation}
    \left( \frac{S_{M \cap M_{i}}}{S_{M_{i}}} < t_{m_1} \right) \lor \left( \frac{S_{M \cap M_{\hat{o}}}}{S_{M_{\hat{o}}}} < t_{m_2} \right) \lor \left( \frac{S_{M \cap M_{\hat{o}}}}{S_{M}} < t_{m_3} \right).
\end{equation}

Moreover, experiments show that SAM2 often produces noisy segmentation when the object is absent in certain frames. To prevent excessive refinement prompts, we first check for such cases. Assuming constant camera speed, if the object was absent in the previous frame and expected to stay out of view, any noisy segmentation is discarded, and the object mask is automatically set to zero.

\subsection{Decomposed Scene Reconstruction}
In the previous section, we outlined the process for obtaining accurate segmentation results for the interacted object in the 2D image. Now, we will focus on how to utilize these 2D segmentation results to construct and optimize our 3D decomposed map.

\noindent{\bf Progressive decomposition.}
Upon detecting a new instance that undergoes interactions, we decouple it from the original 3D map and represent it separately using a dedicated set of Gaussians. This process begins by extracting the object's mask from the current frame and propagating it to the past keyframes using the prompted segmentation method described earlier. With segmentation results available from multiple viewpoints, we project each Gaussian onto the 2D camera plane of these frames based on the estimated camera poses. As shown in~\cref{eq:criteria}, Gaussians $\tilde{g}$ that frequently appear within the mask are considered part of the object, and they are decoupled into an independent set, assigned a new object ID.

\begin{equation}
    \frac{\sum_{f \in \mathcal{F}_{\text{valid}}} \mathbbm{1}\left( P(\tilde{g}) \in M_r \right)}{|\mathcal{F}_{\text{valid}}|} > t_{3d},
    \label{eq:criteria}
\end{equation}
where $\mathcal{F}_{\text{valid}}$ represents the set of keyframes containing a complete observation of the interacted object, and $M_r$ represents the refined mask in these keyframes.

\noindent{\bf Joint optimization.}
Once the camera and object poses are estimated, the next step is to perform global optimization to refine both object decomposition and reconstruction. The optimization is guided by a global objective function, as defined in~\cref{eq:global}, where the scene’s color, depth, and object ID serve as supervision signals with weights $\lambda_p^{map}$, $\lambda_d^{map}$ and $\lambda_{ID}^{map}$.
For each training iteration, keyframes are sampled from the keyframe buffer and trained alongside the current frame. All parameters, including camera poses, object poses, and Gaussian parameters, are jointly optimized, except for object poses outside interaction periods. This joint optimization serves a role similar to bundle adjustment in SLAM, where keyframe replay mitigates catastrophic forgetting and improves the consistency and quality of the reconstructed decomposed map.

\begin{figure*}[ht]
    \centering
    \includegraphics[width=0.9\textwidth]{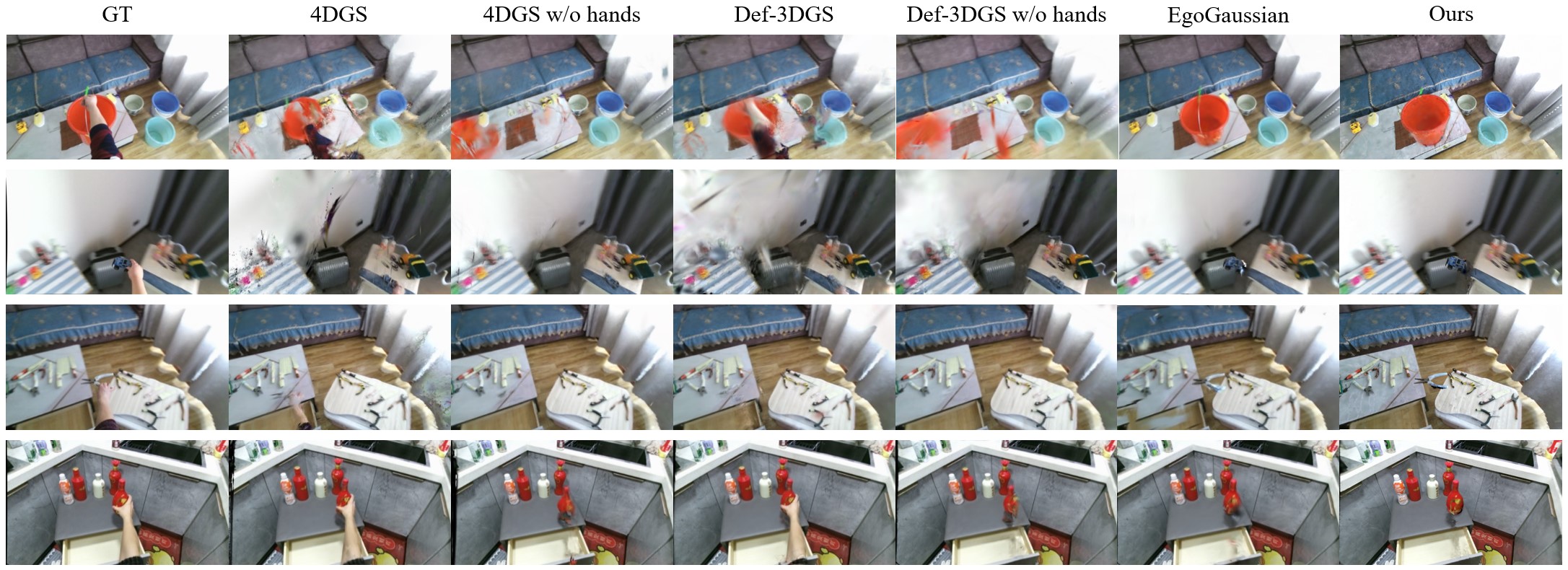}
    \caption{Qualitative comparison of rendering results on HOI4D dataset.}
    \label{fig:psnr}
\end{figure*}

\section{Experiments}
\label{sec:experiments}
Our method aims to fully leverage the proactive interaction information contained in egocentric videos, while simultaneously achieving scene decomposition and reconstruction. To validate the effectiveness of our approach, we conducted evaluations from multiple perspectives. We tested our method on the HOI4D dataset~\cite{liu2022hoi4d}, which is an egocentric dataset containing hand-object interactions. However, since the sequences in HOI4D contain only a small number of interacted objects, they do not adequately demonstrate our method's ability to accurately decompose scenes. Therefore, we propose a more challenging dataset, named the MHOI dataset, which contains ten egocentric RGB-D video sequences, each involving proactive interactions with 3 to 8 different objects. We conducted experiments on both datasets.

\subsection{Experimental Setup}
The experiments are performed on a desktop PC with an Intel i9-12900K CPU and an NVIDIA RTX 4090 GPU.

In our experimental implementation, we set the parameters as follows: loss function coefficients $\lambda_p^{ctrack} = \lambda_p^{otrack} = 0.5$, $\lambda_d^{ctrack} = \lambda_d^{otrack} = \lambda_p^{map} = \lambda_d^{map} = 1.0$, $\lambda_{ID}^{ctrack} = 0.0$, $\lambda_{ID}^{otrack} = \lambda_{ID}^{map} = 2.5$; thresholds for identifying interacted objects: \(t_d = 0.3\), \(t_p = 0.5\); thresholds for segmentation refinement: \(t_{m_1} = t_{m_2} = 0.9\), \(t_{m_3} = 0.7\); threshold for decoupling interacted objects from the background: \(t_{3d} = 0.8\). For each incoming frame, we perform camera tracking and 6-DoF object tracking. Every 10 frames, we conduct joint optimization, and every 30 frames, we store the corresponding frame as a keyframe.

\subsection{Segmentation and Decomposition}

As shown in~\cref{fig:cover}, our method achieves accurate segmentation of interacted objects and progressive scene decomposition. Unlike most current object-level scene reconstruction methods that rely on images captured under static conditions, our approach utilizes proactive interaction to clearly define segmentation granularity and eliminate its ambiguity. ~\cref{fig:gaussian_grouping} clearly illustrates this point. Without leveraging motion information, Gaussian Grouping fails to separate components such as the thermal container and its lid, or the cabinet and its drawers. In contrast, our method accurately decouples each moving unit individually, which is more beneficial for downstream tasks like robotic manipulation.

In our workflow, we use depth inconsistencies as cues to generate prompts, applying SAM2 to obtain the mask of the interacted object and perform mask association. However, depending solely on the results from SAM2 is unreliable. ~\cref{fig:segmentation} illustrates the significant role of our mask refinement by presenting several common failure cases of SAM2. As discussed in~\cref{subsec:mask_refinement}, to address the scenario depicted in~\cref{fig:segmentation} (a), we designed a flexible-length memory bank to mitigate the negative impact of problematic memory features on segmentation, thereby enabling a complete segmentation of the entire kettle. In~\cref{fig:segmentation} (b) and (c), errors are observed in the masks obtained by SAM2's mask association. To correct these segmentation errors, we compare the obtained masks with the inconsistency area and the rendered mask after object tracking, and then add new prompts to refine the results accordingly. ~\cref{fig:segmentation} (d) demonstrates how SAM2 often produces noisy and incorrect segmentation when the object is entirely out of frame. To avoid adding excessive prompts, we determine the object's status based on its 3D position. If the object is projected outside the frame at a given moment, we directly assign a mask with all zeros.

We also conduct quantitative evaluations of decomposition on the HOI4D dataset, using the four sequences shown in~\cref{fig:psnr}. It can be clearly observed in~\cref{tab:mask_comparison} that our rendered masks are more accurate compared to those directly provided by SAM2, especially in Sequence 3, where the interacted object is a structurally complex pair of scissors. Without refinement, SAM2 often segments only the tip of the scissors, failing to capture the entire object.

\begin{table}[h]
\centering
\begin{tabular}{lccccc@{}}
\toprule
\textbf{Method} & \textbf{Seq 1} & \textbf{Seq 2} & \textbf{Seq 3} & \textbf{Seq 4} \\ \midrule

SAM2                 & 0.913 & 0.884 & 0.318 & 0.941 \\
Rendered Mask (Ours) & \textbf{0.925} & \textbf{0.920} & \textbf{0.835} & \textbf{0.947} \\ 
\bottomrule
\end{tabular}
\caption{Comparison of mask quality (mIoU) across sequences on HOI4D dataset.}
\label{tab:mask_comparison}
\end{table}

\vspace{-1.1em}

\begin{table}[h]
\centering
\begin{tabular}{lcccccc}
\toprule
\multirow{2}{*}{\textbf{Method}} &
  \multicolumn{3}{c}{\textbf{HOI4D}} &
  \multicolumn{1}{c}{\textbf{MHOI}} \\ \cmidrule(lr){2-4} \cmidrule(l){5-5} 
  & \textbf{ATE} & \textbf{PSNR (s)} & \textbf{PSNR (d)} & \textbf{ATE} \\ \midrule

Co-SLAM &   0.172      &  17.35    & -- & 0.221 \\
SplaTaM &   0.156      &   18.61   & -- & 0.293 \\
NeuDySLAM & 0.094 & 25.15 & -- & 0.189 \\
Ours       & \textbf{0.076} & \textbf{29.12} & \textbf{27.58} & \textbf{0.093} \\ 
\bottomrule
\end{tabular}

\caption{Comparison with recent SLAM methods in terms of camera localization accuracy (ATE [m]) and rendering quality. Unlike NeuDySLAM and other dynamic SLAM methods that only reconstruct the static scene, our method also reconstructs the interacted objects that are in motion.}
\label{tab:hoi4d_mhoi_comparison}
\end{table}

% \begin{table}[h]
% \centering
% \setlength\tabcolsep{2pt} %
% \resizebox{0.95\linewidth}{!}{%
% \begin{tabular}{lccccccccc@{}}
% \toprule
% \multirow{2}{*}{\textbf{Method}} &
%   \multicolumn{3}{c}{\textbf{Static}} &
%   \multicolumn{1}{l}{} &
%   \multicolumn{3}{c}{\textbf{Dynamic}} &
%   \multirow{2}{*}{\textbf{Iterations}} \\ \cmidrule(lr){2-4} \cmidrule(l){6-8}
%   &
%   \multicolumn{1}{l}{SSIM $\uparrow$} &
%   \multicolumn{1}{l}{PSNR $\uparrow$} &
%   \multicolumn{1}{l}{LPIPS $\downarrow$} &
%   \multicolumn{1}{l}{} &
%   \multicolumn{1}{l}{SSIM $\uparrow$} &
%   \multicolumn{1}{l}{PSNR $\uparrow$} &
%   \multicolumn{1}{l}{LPIPS $\downarrow$} &
%   \\ \midrule
% 4DGS~\cite{wu20244d}           & 0.88 & 25.33 & 0.13 & & 0.89 & 25.34 & 0.13 & 30k  \\
% 4DGS w/o hands & \uline{0.94} & 28.69 & \textbf{0.08} & & \uline{0.94} & 27.33 & \uline{0.10} & 30k  \\
% Def-3DGS~\cite{yang2024deformable}     & 0.90 & 25.85 & \uline{0.11} & & 0.90 & 25.71 & 0.12 & 30k  \\
% Def-3DGS w/o hands & \uline{0.94} & 28.09 & \textbf{0.08} & & \uline{0.94} & 26.92 & \uline{0.10} & 30k  \\
% EgoGaussian~\cite{Zhang2025_3dv}            & \textbf{0.96} & \textbf{30.99} & \textbf{0.08} & & \textbf{0.95} & \textbf{30.33} & \textbf{0.09} & 30k \\
% \midrule
% Ours           &    \textbf{0.96}  & \uline{29.12} & \textbf{0.08} &  & 0.92  & \uline{27.58} & \uline{0.10} & 50+50 \\
% \bottomrule
% \end{tabular}%
% }
% \caption{Quantitative comparison of novel view synthesis results with 4DGS, Def-3DGS, and EgoGaussian.}
% \label{tab:hoi4d_results}
% \end{table}

\begin{table}[h]
\centering
\setlength\tabcolsep{2pt} % 增加列间距
\resizebox{0.95\linewidth}{!}{%
\begin{tabular}{llccccccccc} % Offline 列变宽
\toprule
\multirow{2}{*}{} & 
\multirow{2}{*}{\textbf{Method}} &
  \multicolumn{3}{c}{\textbf{Static}} &
  & 
  \multicolumn{3}{c}{\textbf{Dynamic}} &
  \multirow{2}{*}{\textbf{Iterations}} \\ 
\cmidrule(lr){3-5} \cmidrule(lr){7-9}
 & & SSIM $\uparrow$ & PSNR $\uparrow$ & LPIPS $\downarrow$ &  
   & SSIM $\uparrow$ & PSNR $\uparrow$ & LPIPS $\downarrow$ &  \\ 
\midrule
\multirow{5}{*}{\centering\rotatebox{90}{\textbf{\hspace{5pt} Offline}}} 
& 4DGS~\cite{wu20244d}           & 0.88 & 25.33 & 0.13 & & 0.89 & 25.34 & 0.13 & 30k  \\
& 4DGS w/o hands & \uline{0.94} & 28.69 & \textbf{0.08} & & \uline{0.94} & 27.33 & \uline{0.10} & 30k  \\
& Def-3DGS~\cite{yang2024deformable}     & 0.90 & 25.85 & \uline{0.11} & & 0.90 & 25.71 & 0.12 & 30k  \\
& Def-3DGS w/o hands & \uline{0.94} & 28.09 & \textbf{0.08} & & \uline{0.94} & 26.92 & \uline{0.10} & 30k  \\
& EgoGaussian~\cite{Zhang2025_3dv}            & \textbf{0.96} & \textbf{30.99} & \textbf{0.08} & & \textbf{0.95} & \textbf{30.33} & \textbf{0.09} & 30k \\
\midrule
& Ours           &    \textbf{0.96}  & \uline{29.12} & \textbf{0.08} &  & 0.92  & \uline{27.58} & \uline{0.10} & $\sim$4k \\
\bottomrule
\end{tabular}%
}
\caption{Quantitative comparison of novel view synthesis results with 4DGS, Def-3DGS, and EgoGaussian.}
\label{tab:hoi4d_results}
\end{table}

\vspace{-0.3em}

\subsection{Camera Tracking and Scene Reconstruction}

\noindent{\bf Camera tracking.}
We compared our method with three recent SLAM approaches: Co-SLAM and SplaTaM, which are based on a static scene assumption, and NeuDySLAM~\cite{li2024learn}, the state-of-the-art NeRF-based dynamic SLAM method. As shown in~\cref{tab:hoi4d_mhoi_comparison}, our approach achieves superior localization accuracy on both datasets. This difference is especially pronounced on the MHOI dataset, which contains multiple interacted objects. For NeuDySLAM, We attribute this to the fact that it masks out all objects that have experienced motion when optimizing the camera pose, whereas our method only masks out objects that are currently moving due to interaction. Therefore, when there are numerous interacted objects, NeuDySLAM discards too many useful features, resulting in a decrease in accuracy due to the lack of available features.

\noindent{\bf Scene reconstruction.} 
We use novel view synthesis results to evaluate the quality of the reconstructed map. \cref{fig:cover} and~\cref{fig:psnr} show the qualitative results of our method on the MHOI and HOI4D datasets, respectively. Our method, as shown, produces high-quality rendering for both the background and interacted object parts. Moreover, it also achieves accurate object tracking\footnote{Visualization results are provided in the supplementary material}. These results demonstrate the effectiveness of our decomposed scene reconstruction. Notably, our method also possesses the capability to accurately reconstruct articulated objects and estimate their kinematics.\footnotemark[1].

We also perform quantitative evaluations on the HOI4D dataset, using the experimental settings from EgoGaussian~\cite{Zhang2025_3dv} and including some results in~\cref{tab:hoi4d_results} based on the original EgoGaussian experiment. Specifically, four sequences are selected to evaluate the rendering quality of both dynamic and static parts separately. We compare our method against 4DGS~\cite{wu20244d}, Def-3DGS~\cite{yang2024deformable}, and EgoGaussian. The first two methods are designed for non-rigid motion, whereas EgoGaussian and our method are targeted at scenes involving hand interactions with rigid objects. For a fairer comparison, modifications are made to the other two methods so that hands can be masked out.

As to the quantitative results, our method outperforms 4DGS and Def-3DGS while achieving metrics close to EgoGaussian, yet requires only a small fraction of the optimization iterations used by the other methods. Additionally, they rely on accurate camera poses as input, whereas our method performs its own camera tracking as a SLAM system.

\section{Conclusion}
In this paper, we introduce the task of proactive scene decomposition and reconstruction, which aims to adaptively decompose and reconstruct dynamic environments on the fly based on human-object interactions. To tackle the problem, we propose an online dynamic SLAM system that iteratively refines the map representation and the corresponding composition through interaction cues. Our approach is verified through experiments in camera pose estimation, object decomposition, and scene reconstruction, achieving high-quality and accurate modeling of dynamic environments. The results confirm the effectiveness of our system in capturing and representing the dynamic nature of the environment through proactive interactions.

\clearpage

\section*{Acknowledgement}
We gratefully acknowledge the anonymous reviewers and AC for their valuable comments and suggestions. This work is supported by NSFC (U22A2061, 62176010) and 230601GP0004.

{
    \small
    \bibliographystyle{ieeenat_fullname}
    \bibliography{main}
}

\clearpage
\setcounter{page}{1}
\maketitlesupplementary

\section{Supplementary Results}
\label{sec:supplementary}

\subsection{Interacted Object Trajectory Visualization}
\begin{figure*}[h!]
    \centering
    \includegraphics[width=\textwidth]{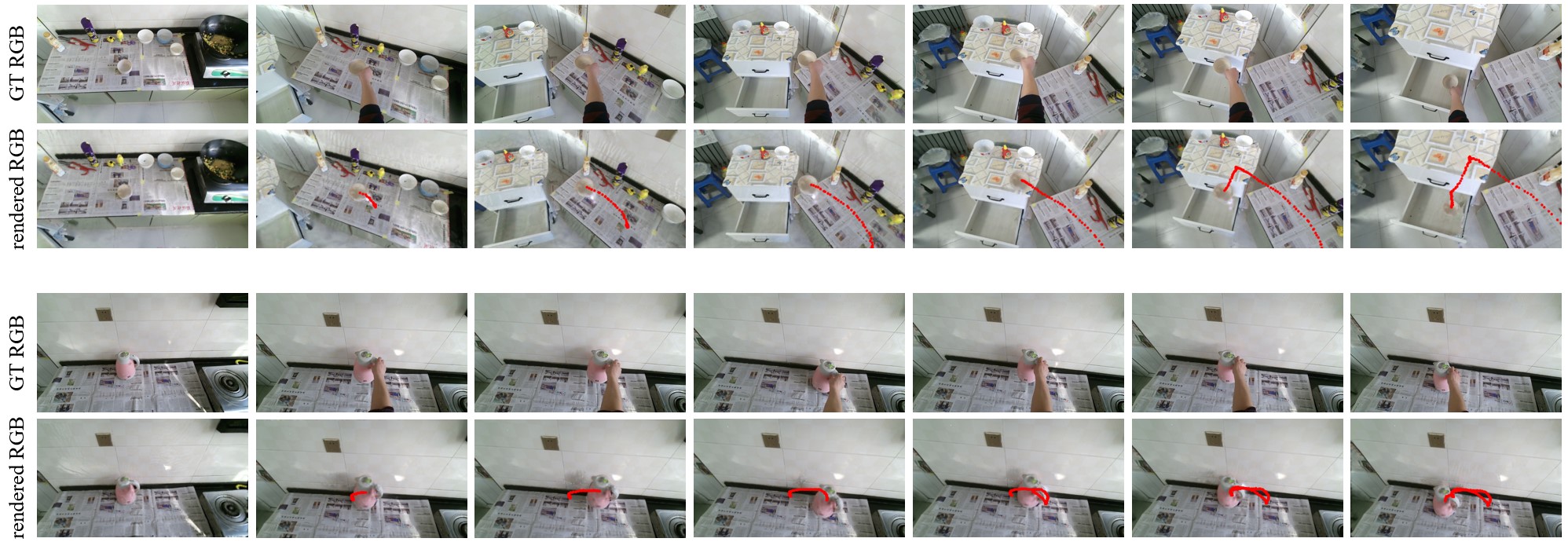}
    \caption{Visualization of the interacted object trajectory in HOI4D dataset.}
    \label{fig:object_trajectory}
\end{figure*}

To visually demonstrate the accuracy of our method in object pose tracking, we visualize the trajectory of the interacted object. As shown in~\cref{fig:object_trajectory}, the red dots in the second row of images represent the position of the object at each time step. From this visualization, it is evident that our method accurately estimates the object motion, which also indicates the reliability of our decomposition and scene reconstruction.

\subsection{Progressive Decomposition and Online Reconstruction}
\begin{figure*}[h]
    \centering
    \includegraphics[width=\textwidth]{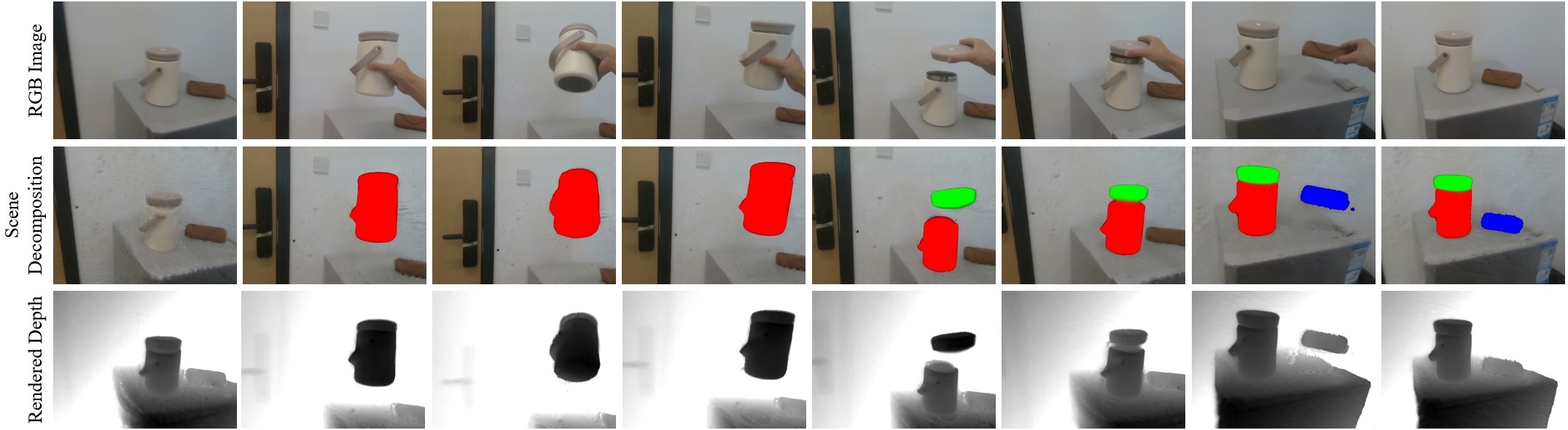}
    \caption{Qualitative results of progressive scene decomposition and reconstruction via proactive interaction.}
    \label{fig:supp_teaser}
\end{figure*}

Our method progressively achieves scene decomposition and reconstruction through proactive user interaction. Here, we provide~\cref{fig:supp_teaser} as an extra qualitative demonstration, which shows that our resulting map achieves accurate decomposition and high-quality rendering.

\begin{figure}[ht]
    \centering
    \includegraphics[width=0.48\textwidth]{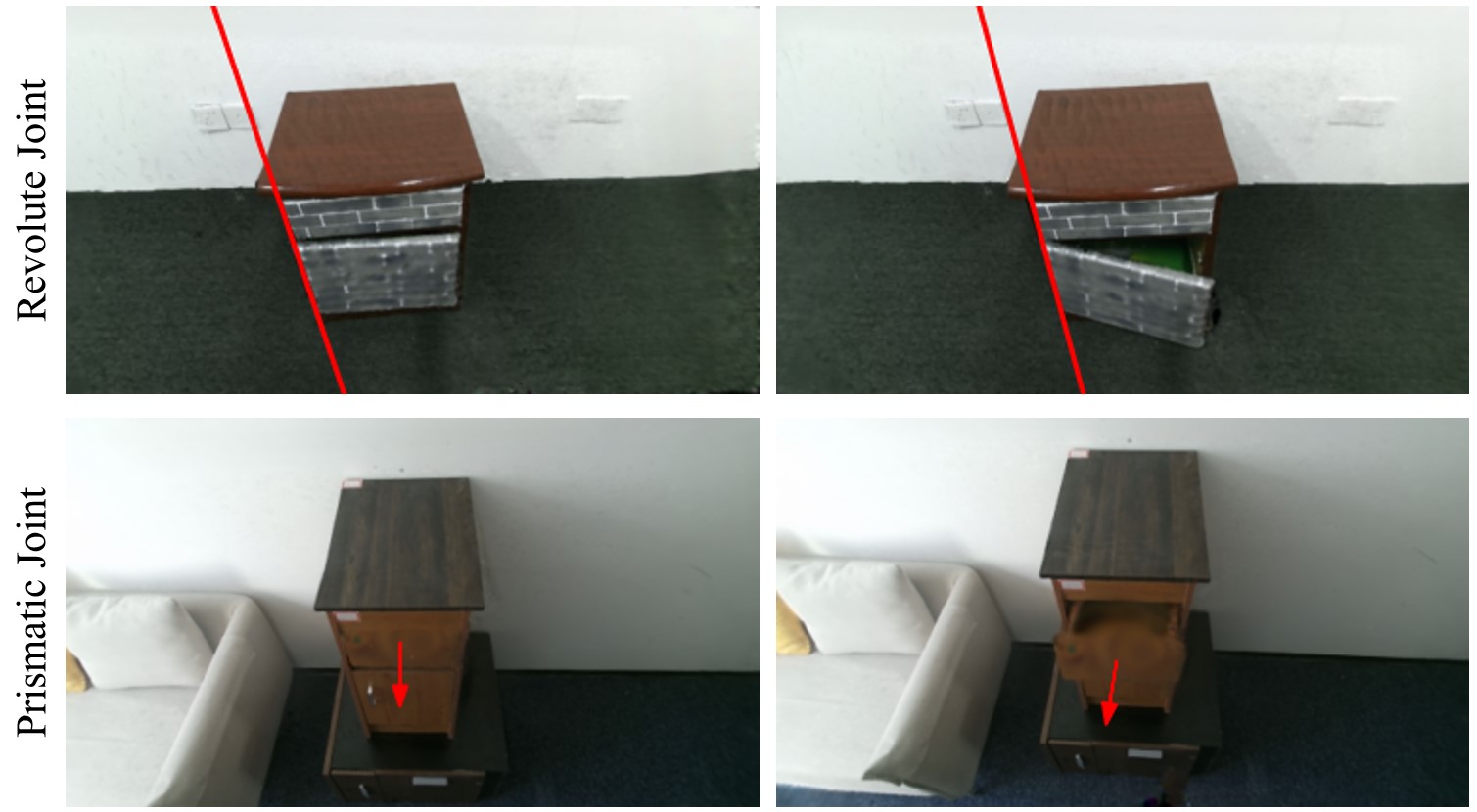}
    \caption{Visualization of the revolute axis and prismatic joint of articulated objects.}
    \label{fig:articulation}
\end{figure}

\subsection{Run-time}
We evaluate the average run-time of our system on the four sequences from HOI4D dataset, with the detailed timing of each module presented in~\cref{tab:module_timing}. Our approach achieves a good balance between system performance and run-time, ensuring accurate decomposition and reconstruction while demonstrating significant efficiency advantages over offline methods.

\begin{table}[h]
\centering
\setlength\tabcolsep{2pt} %
\resizebox{0.98\linewidth}{!}{%
\begin{tabular}{lcccc}
\toprule
 &
  \makecell{\textbf{Object} \\ \textbf{Segmentation}} &
  \makecell{\textbf{Camera} \\ \textbf{Tracking}} &
  \makecell{\textbf{Object} \\ \textbf{Tracking}} &
  \makecell{\textbf{Joint} \\ \textbf{Optimization}} \\ \midrule

Time (ms) & 429 & 371 & 344 & 493 \\ 
\bottomrule
\end{tabular}
}
\caption{Per-frame time consumption of modules in our system.}
\label{tab:module_timing}
\end{table}

\subsection{Object Pose Evaluation}
We also evaluate 6-DoF object pose tracking results on the HOI4D dataset (\cref{tab:object_pose}). For evaluation, we use the Area Under the Curve (AUC) of the ADD and ADDS metrics, which are defined as follows:

\[
    \text{ADD} = \frac{1}{m} \sum_{v \in \mathcal{O}} \| (R v + T) - (R^* v + T^*) \|
\]

\[
    \text{ADDS} = \frac{1}{m} \sum_{v_1 \in \mathcal{O}} \min_{v_2 \in \mathcal{O}} \| (R v_1 + T) - (R^* v_2 + T^*) \|
\]
where $\mathcal{O}$ is the set of object vertices, $R, T$ and $R^*, T^*$ are the predicted and ground truth rotation and translation.

\begin{table}[h]
\centering
\setlength\tabcolsep{22pt}
\begin{tabular}{lcc}
\toprule
\textbf{Metric} &
  \textbf{ADD-S} &
  \textbf{ADD} \\ \midrule

Mean (\%) & 86.44 & 80.35 \\ 
\bottomrule
\end{tabular}

\caption{The ADD and ADD-S metrics are reported as AUC percentages (0 to 0.1 m) on the tested sequences in HOI4D dataset.}
\label{tab:object_pose}
\end{table}

\subsection{Ablation Study}
In Sec. 4.3, we discuss three strategies for mask refinement: A) flexible memory bank, B) absence check, and C) inter-frame inconsistency check. Here, we illustrate their effectiveness through ablation studies in~\cref{tab:ablation1}. We also include an ablation study on hyperparameters. The results in~\cref{tab:ablation2} and ~\cref{tab:ablation3} demonstrate the appropriateness of the hyperparameter values we chose for mask refinement and dynamic object identification.

\begin{table}[ht]
\centering
\resizebox{0.95\linewidth}{!}{%
\begin{tabular}{lcccc}
\hline
  & w/o A & w/o B & w/o C & Ours \\
\hline
Refined mask (mIoU) & 0.904 & 0.885 & 0.723 & 0.917 \\
\hline
\end{tabular}
}
\vspace{-0.5em}
\caption{The impact of different strategies on mask refinement.}

\label{tab:ablation1}
\end{table}

\begin{table}[h]
\centering
\begin{minipage}{0.45\linewidth} % 这里调整宽度
\centering
\resizebox{1.2\linewidth}{!}{%
\begin{tabular}{cccc}
\hline
\diagbox{\(t_{m_1, m_2}\)}{\(t_{m_3}\)} & 0.6 & 0.7 & 0.8 \\ \hline
  0.8 & 0.861 & 0.912 & 0.912 \\ \hline
     0.9 & 0.898 & 0.917 & 0.915 \\ \hline
        0.95 & 0.903 & 0.917 & 0.910 \\ \hline
\end{tabular}%
}
\caption{Ablation on hyperparameters \(t_{m_1}, t_{m_2}, t_{m_3}\).}
\label{tab:ablation2}
\end{minipage} \hspace{0.65cm}
\begin{minipage}{0.45\linewidth} % 这里调整宽度
\centering
\resizebox{0.93\linewidth}{!}{%
\begin{tabular}{cccc}
\hline
\diagbox{\(t_d\)}{\(t_p\)} & 0.4 & 0.5 & 0.6 \\ \hline
 0.2 & 0.836 & 0.895 & 0.908 \\ \hline
 0.3 & 0.902 & 0.917 & 0.883 \\ \hline
 0.4 & 0.917 & 0.870 & 0.745 \\ \hline
\end{tabular}%
}
\caption{Ablation on hyperparameters \(t_{d}, t_{p}\).}
\label{tab:ablation3}
\end{minipage}
\end{table}

\section{Decomposition and Reconstruction of Articulated Objects}

We assume that object motion follows a 6-DoF rigid body motion, allowing our method to model articulated objects as well. Specifically, after completing the decomposition and reconstruction using an input sequence, we examine each object individually. As shown in~\cref{fig:articulation}, if the object's movement is restricted to translation or rotation only, which corresponds to a prismatic joint or a revolute joint respectively, we classify it as an articulated object and proceed to calculate its kinematics. The detailed evaluation process is as follows:

\noindent{\bf Pure Translation (prismatic joint)} occurs when the rigid body moves along a single direction without rotation. To identify such motion, two key properties are evaluated: (1) the rotation angles derived from the quaternions should be negligible, and (2) the translation vectors should exhibit approximate collinearity.

The quaternion \( \mathbf{q}_t = [w_t, x_t, y_t, z_t] \) represents the rotation at time \( t \). The corresponding rotation angle \( \theta_t \) is computed as:
\[
\theta_t = 2 \arccos(w_t),
\]
where \( w_t \) is the scalar component of the quaternion. For pure translation, \( \theta_t \) should be close to zero within a predefined threshold \( \epsilon_\text{rot} \):
\[
|\theta_t| < \epsilon_\text{rot}, \quad \forall t \in \{1, \dots, N\}.
\]

Additionally, the translation vectors \( \mathbf{T} \) should align approximately along a single direction. This property can be evaluated by first centralizing the translations to remove the mean displacement:
\[
\mathbf{T}_\text{centered} = \mathbf{T} - \bar{\mathbf{T}}, \quad \bar{\mathbf{T}} = \frac{1}{N} \sum_{t=1}^N \mathbf{T}_t.
\]
The covariance matrix of the centralized translations is then computed as:
\[
\mathbf{C} = \frac{1}{N} \mathbf{T}_\text{centered}^\top \mathbf{T}_\text{centered}.
\]
The eigenvalues of \( \mathbf{C} \) describe the variance of the translations along the principal axes. For approximately collinear translations, the largest eigenvalue \( \lambda_1 \) should dominate, while the remaining eigenvalues \( \lambda_2, \lambda_3 \) should remain small relative to \( \lambda_1 \). Specifically, we define the collinearity condition as:
\[
\frac{\lambda_2 + \lambda_3}{\lambda_1} < \epsilon_\text{eig}.
\]

Thus, a motion is classified as pure translation if \( |\theta_t| < \epsilon_\text{rot} \) for all \( t \) and the eigenvalue ratio satisfies \( (\lambda_2 + \lambda_3) / \lambda_1 < \epsilon_\text{eig} \).

\noindent{\bf Pure Rotation Around a Fixed Axis (revolute joint)} occurs when the rigid body moves about an axis without any significant translational displacement along the axis. To identify such motion, two conditions are evaluated: (1) the consistency of the rotation axis across all time steps, and (2) the translational displacement parallel to the axis must be negligible.

The rotation axis at time \( t \) can be extracted from the quaternion \( \mathbf{q}_t = [w_t, x_t, y_t, z_t] \) as:
\[
\mathbf{v}_t = \frac{(x_t, y_t, z_t)}{\sqrt{x_t^2 + y_t^2 + z_t^2}},
\]
where \( x_t, y_t, z_t \) are the vector components of \( \mathbf{q}_t \). The mean rotation axis is computed and normalized as:
\[
\mathbf{v} = \frac{\sum_{t=1}^N \mathbf{v}_t}{\left\|\sum_{t=1}^N \mathbf{v}_t \right\|}.
\]
The consistency of \( \mathbf{v}_t \) is evaluated using the angular deviation:
\[
\Delta\theta_t = \arccos(\mathbf{v}_t \cdot \mathbf{v}),
\]

In addition to the axis consistency, the translational displacement parallel to the axis is evaluated. For each time step \( t \), the translation vector \( \mathbf{T}_t \) is decomposed as:
\[
\mathbf{T}_t^\parallel = (\mathbf{T}_t \cdot \mathbf{v}) \mathbf{v}, \quad \mathbf{T}_t^\perp = \mathbf{T}_t - \mathbf{T}_t^\parallel.
\]
Here, \( \mathbf{T}_t^\parallel \) represents the displacement along the axis. The motion is classified as pure rotation around a fixed axis if and only if both conditions are satisfied:
\[
\max_t |\Delta\theta_t| < \epsilon_\text{axis} \quad \text{and} \quad \max_t \|\mathbf{T}_t^\parallel\| < \epsilon_\text{trans}.
\]

\section{Potential Downstream Applications}
Our method eliminates the ambiguity of segmentation granularity in a proactive manner. It incorporates priors from interactions and introduces structured representations through object decomposition. Additionally, it achieves photorealistic rendering results via Gaussian splatting. Consequently, compared to other approaches, our method delivers a structured scene map with accurate decomposition and high-fidelity rendering. As a result, the generated map can be effectively applied to zero-shot mobile manipulation, robot learning, photorealistic simulation, and immersive VR and AR applications.

\section{Limitations and Future Work}
Our method has some limitations. Currently, object tracking in our approach is restricted to rigid body motion, which limits its ability to model more flexible dynamics. Extending the method to handle non-rigid objects, such as deformable surfaces, would improve its applicability to more complex real-world interactions.

Additionally, while the core insight—resolving decomposition ambiguity through proactive interaction—is generalizable, the reliance on hand guidance makes it more suited for egocentric views. Future work could explore alternative guidance mechanisms to expand its use beyond egocentric settings and adapt to a broader range of applications.

\end{document}